\documentclass[b5paper,10pt,abstractoff,DIV=calc,headings=normal]{scrartcl}
\usepackage[english]{babel}
\usepackage{lmodern}
\usepackage{microtype}
\usepackage{amsmath,amssymb,amsthm,amsfonts,graphicx,etoolbox,scrlayer-scrpage}
\usepackage[noblocks]{authblk}
\usepackage{url}

\usepackage{subcaption}
\usepackage{caption}

\makeatletter
\patchcmd{\@maketitle}{\huge}{\Large}{}{}
\patchcmd{\abstract}{\quotation}{}{}{}
\AtBeginEnvironment{abstract}{\noindent\ignorespaces}
\AtEndEnvironment{abstract}{\par\mbox{}}
\newcommand{\shortauthor}{}
\newcommand{\shorttitle}{\@title}
\makeatother
\setkomafont{author}{\small}
\setkomafont{title}{\rmfamily\bfseries}
\setkomafont{disposition}{\rmfamily\bfseries}

\newcommand{\acknowledgements}{\par\mbox{}\par\noindent\textbf{Acknowledgements: }}
\newcommand{\keywords}[1]{\par\noindent\textbf{Keywords: }#1}
\rehead[]{\shortauthor}\lohead[]{\shorttitle}\lehead[]{~}\rohead[]{~}
\theoremstyle{plain}

\theoremstyle{definition}

\theoremstyle{remark}

\renewenvironment{abstract}{\bigskip\noindent\begin{minipage}{\textwidth}\setlength{\parindent}{15pt}\paragraph{Abstract:}}{\end{minipage}}

\newcommand{\dd}{\mathop{}\! \mathrm{d}}
\newcommand{\argmax}{\operatorname*{\arg\max}}

\usepackage{algpseudocode}
\usepackage{algorithm}
\algrenewcommand\algorithmicindent{0.5em}%

\begin{document}


\renewcommand{\shortauthor}{~}

\title{Adversarial attacks against Bayesian forecasting dynamic models}

\author{Roi Naveiro\thanks{Corresponding author: roi.naveiro@icmat.com}}
\affil{Institute of Mathematical Sciences (ICMAT-CSIC), Madrid, Spain.}

\date{\vspace{-5ex}}
\maketitle


\begin{abstract}

The last decade has seen the rise of Adversarial Machine Learning (AML). This discipline studies how to manipulate data to fool inference engines, and how to protect those systems against such manipulation attacks. Extensive work on attacks against regression and classification systems is available, while little attention has been paid to attacks against time series forecasting systems. In this paper, we propose a decision analysis based attacking strategy that could be utilized against Bayesian forecasting dynamic models. 

\end{abstract}

\keywords{Bayesian forecasting, Adversarial machine learning, Bayesian model monitoring. }


\section{Introduction}

Machine learning (ML) applications have experienced an impressive growth over the last decade. However, the ever increasing adoption of ML methodologies has revealed important security issues. Among these, vulnerabilities to adversarial examples \cite{goodfellow2014explaining}, intentionally manipulated data instances targeted at fooling ML algorithms, are especially important. 
Examples abound. For instance, it is relatively easy to fool a spam detector simply misspelling spam words. Obfuscation of malware code can make it seem legitimate. Simply adding stickers to a stop sign could make an autonomous vehicle classify it as a merge sign. Consequences could be catastrophic.
In contexts in which ML systems are susceptible of being attacked, algorithms should acknowledge the presence of possible adversaries and be trained in such a way that they are robust against their potential data manipulations. This is the main goal of AML.
As recently pointed out in \cite{naveiro2018adversarial}, having relevant models of how an adversary might modify input data to a learning system is key to guarantee protection against adversarial attacks. 
Indeed, probabilistic approaches to AML as \cite{rios2020perspectives}, suggest that it is key to be able to sample from the distribution of original unperturbed data, given the observed (possibly perturbed) one. 
To make that sampling feasible, it is necessary to have probabilistic models of how an adversary might modify a data instance. 
Those models must take into account the existing uncertainties: that of adversary about the learning system, and the one the modeller has about the adversary.

In the last few years, several attacking models to classification and regression systems have been released. 
Probably, the most popular ones target computer vision, \cite{akhtar2018threat}, 
but others have appeared in fields like fake news detection \cite{zhou2019fake}. 
However, much less attention has been paid to attacks to time series forecasting models. In this paper, we put ourselves in the shoes of an adversary willing to manipulate input data to a time series forecasting system in order to drive predictions to a target of his interest. This is a crucial step prior to developing robust defence mechanisms.





\section{Attacks against Bayesian forecasting dynamic models}\label{sec:poi_bay}


Previously proposed attacks against time series forecasting models \cite{alfeld2016data} focused on attacking simple auto-regressive (AR) models. Their idea is straightforward: an adversary is interested in driving the predictions made by an AR model during a fixed time window towards a target of his interest. To do so, he deliberately modifies the observations received by the AR system prior to the onset of the time window. 
The adversary is assumed to have complete knowledge about the parameters governing the AR model and the original data that would be fed into the system.
In this case, given a candidate data manipulation, the attacker could compute the vector of predictions that the model would yield. A data manipulation is then selected to minimize the distance between the predictions produced and the target predictions, subject to some constraints about the size of the data manipulation reflecting a limited budget or the fact that the adversary wants to avoid being detected.

In this paper, our main purpose is to propose attacking strategies that target Bayesian forecasting dynamic models \cite{west2006bayesian}. Rather than extending previous attacks to target these models, we build a novel, more realistic, attacking strategy based on decision analysis ideas. Let us start with the setting.
We consider two agents: a Defender ($D$, she), which is implementing a Bayesian forecasting dynamic model to aid her in making some decision, and an attacker ($A$, he), manipulating the data that $D$ receives, in order to modify such a decision. As a running example, consider that $D$ is an ad company monitoring traffic flow into one of its webpages, using a conditionally Poisson dynamic model \cite{chen2018scalable}. At some point, based on predictions for the next few days, $D$ must decide whether to place an ad in the monitored node or not. In turn, $A$ modifies the input data of the model by producing fake connections to the website in order to induce $D$ to make a wrong decision. For instance, $A$ could be interested in making $D$ spend resources on placing the ad, when, from a decision analysis point of view, was not convenient. 

To fix ideas, say $D$ is using a Bayesian dynamic model to forecast a quantity $y_t$ of interest. Inferences about this quantity are updated sequentially as new data are observed, with $\mathcal{D}_t$ representing all information available at time $t$.\footnote{$\mathcal{D}_t$ is recursively defined through $\mathcal{D}_t = \mathcal{D}_{t-1} \cup \lbrace y_t \rbrace$.}
When $t=\alpha$, the Defender must make a decision based on her forecasts from time $\alpha+1$ until time $\beta$. From a decision analysis perspective, her optimal decision should maximize her posterior predictive utility
\begin{equation*}
    d^* = \argmax_{d} \Psi(d \vert \mathcal{D}_\alpha) = \argmax_{d} \int u(d, y_{\alpha+1:\beta} ) p( y_{\alpha+1:\beta} \vert \mathcal{D}_\alpha) \dd y_{\alpha+1:\beta},
\end{equation*}
where $u(d, y_{\alpha+1:\beta} )$ is the utility perceived from making decision $d$ when future data is $y_{\alpha+1:\beta}$ and $p( y_{\alpha+1:\beta} \vert \mathcal{D}_\alpha)$ is the posterior predictive distribution over the next $\beta - \alpha$ periods at time $\alpha$. 

Consider now $A$'s problem. We study the case of a very powerful attacker: $A$ has knowledge about $D$'s utility, her probability model, and also the observations that the defender receives from time 0 until time $\alpha$. For instance, $A$ could be an insider within $D$'s company. Assume that $A$ is interested in modifying the observations to be received by $D$ from time $t=\alpha-h$ until time $t=\alpha$, so that, $D$ will update her model with contaminated data $\tilde{y}_{\alpha-h}, \dots, \tilde{y}_{\alpha}$ rather than with clean data $y_{\alpha-h}, \dots, y_{\alpha}$. We denote the attacked data until time $\alpha$ as $\tilde{\mathcal{D}}_\alpha$. Through these modified data, $A$'s goal is to make the Defender decide $d_A$ instead of $d^*$. A main difference between our approach and previous ones is that earlier work on attacks to time series forecasting systems considered that the attacker modifies data to drive the predictions produced by the defender towards a certain target. Instead, we consider that a more encompassing objective to the attacker is to make the defender decide something inconvenient. 

Clearly, not every data manipulation that induces decision $d_A$  is equally interesting for $A$. He might have limited resources and, more importantly, he would typically want to avoid being detected. In previous work, this later goal was formalized limiting the size of the perturbation $\Vert \tilde{y}_{\alpha-h:\alpha} - y_{\alpha-h:\alpha} \Vert$, under certain norm. This would yield the following optimization problem to be solved by the attacker when looking for optimal manipulations
\begin{equation} \label{eq:opt1}
\begin{aligned}
& \min_{\tilde{y}_{\alpha-h:\alpha}} 
& &  \Vert \tilde{y}_{\alpha-h:\alpha} - y_{\alpha-h:\alpha} \Vert  & & \text{s.t.} & & \Psi(d_A \vert \tilde{\mathcal{D}}_\alpha) > \Psi(d \vert \tilde{\mathcal{D}}_\alpha) & & \forall d.
\end{aligned}
\end{equation}
Resource limitations could be incorporated as additional constraints.

We argue that this way of formalizing the goal of wanting to avoid detection, could be inconvenient in several scenarios: even slight changes in the data, if they happen always in the same direction, could induce structural changes in time series that can be easily detected by an appropriate monitoring strategy. 
Imagine that $D$ is indeed monitoring the predictive performance of her Bayesian forecasting model using the strategy described in \cite{west1986bayesian}. The main idea is to compare, at each time, the predictive performance of the current model, with that of an alternative model using the \textit{local Bayes factor}
\begin{equation*}
    H_t = \frac{p(y_t \vert \mathcal{D}_{t-1})}{p_A(y_t \vert \mathcal{D}_{t-1})}.
\end{equation*}
A small $H_t$ indicates low predictive performance at time $t$, and, thus, $y_t$ is considered discrepant. However, in order to be able to detect not only single discrepant points but also structural changes, \cite{west1986bayesian} proposes looking for the most discrepant group of recent, consecutive observations, which entails calculating at each time the \textit{minimum cumulative Bayes factor} 
\begin{equation*}
     V_t = \min_{1\leq k \leq t} H_t H_{t-1} \dots H_{t-k+1}.
\end{equation*}
This can be sequentially computed as $V_t = H_t \min [1, V_{t-1}]$. The basic diagnostic mode of operation of $D$ would be to accept the current model as satisfactory unless $V_t$ falls below some threshold value $\gamma$. This monitoring strategy has been proved to be useful to detect  model failures happening due to outliers or structural changes.

Thus, to avoid detection, data manipulations should not produce significant decrease in the the minimum of the cumulative Bayes factors. Otherwise, they could trigger the monitor. 
We therefore propose looking for attacks, solving the problem
\begin{equation} \label{eq:opt2}
\begin{aligned}
& \max_{\tilde{y}_{\alpha-h:\alpha}} 
& &  \min  \tilde{V}_{\alpha-h:\alpha} & & \text{s.t.} & &  \Psi(d_A \vert \tilde{\mathcal{D}}_\alpha) > \Psi(d \vert \tilde{\mathcal{D}}_\alpha) & & \forall d, 
\end{aligned}
\end{equation}
where $\tilde{V}_t$ is the minimum of the cumulative Bayes factors at time $t$, under attacked data. We believe that this formalization of the concept of \textit{avoiding detection} is more coherent than that in \eqref{eq:opt1}, as it modifies data trying to mimic the predictive behaviour of the model being used by $D$, and thus produces tainted data points that are aligned with $D$'s beliefs.  

Of course, solving problems \eqref{eq:opt1} and \eqref{eq:opt2} exactly is unfeasible, and we must use heuristic methods to approximate the optimal solution. In the next Section, problem \eqref{eq:opt2} is approximately solved using simulated annealing, Algorithm \ref{alg:SA}.

\begin{algorithm}[!htbp]
\caption{Simulated Annealing for approximating optimal perturbation} 
\label{alg:SA}
\hspace*{\algorithmicindent} \textbf{Input: } 
 Min. inverse temperature ($\gamma_{min}$); Max. inverse temperature ($\gamma_{max}$); Rate for inverse temperature increase ($\Delta$); Initial attack; Dynamic model; Proposal function.
\begin{algorithmic}[1]
\State Initialize attack $\tilde{y}_{\alpha-h:\alpha}$ in feasible region
\State Update dynamic model with attacked data and compute $S =  \min  \tilde{V}_{\alpha-h:\alpha}$
\State Set $\tilde{y}^*_{\alpha-h:\alpha} = \tilde{y}_{\alpha-h:\alpha}$ and $S^* = S$
\State Set $\gamma = \gamma_{min}$
\While {$\gamma < \gamma_{max}$}
\State Propose attack $\tilde{y}'_{\alpha-h:\alpha}$ in feasible region.
\State Update dynamic model with such data and compute $ S' = \min  \tilde{V}'_{\alpha-h:\alpha}$
\State If $S' > S$, update best attack
        \begin{eqnarray*}
            y^*_{\alpha-h:\alpha} &=& \tilde{y}'_{\alpha-h:\alpha}\\
            S^* &=& S'
        \end{eqnarray*}
\State With probability $\exp(-\gamma(S - S') )$,  update current attack
        \begin{eqnarray*}
            \tilde{y}_{\alpha-h:\alpha} &=& \tilde{y}'_{\alpha-h:\alpha}\\
            S &=& S'
        \end{eqnarray*}
        
\State Increase $\gamma$ (decrease temperature): $\gamma = \gamma \cdot \Delta$
\EndWhile
\State \textbf{Return:} $ y^*_{\alpha-h:\alpha}$
\end{algorithmic}
\end{algorithm}




\section{Experiments}

\subsection{The add company problem}
\begin{figure}[h!] 
        \centering
        \captionsetup{justification=centering}
        \begin{subfigure}[b]{0.49\textwidth}
            \centering
            \includegraphics[width=1.0\textwidth]{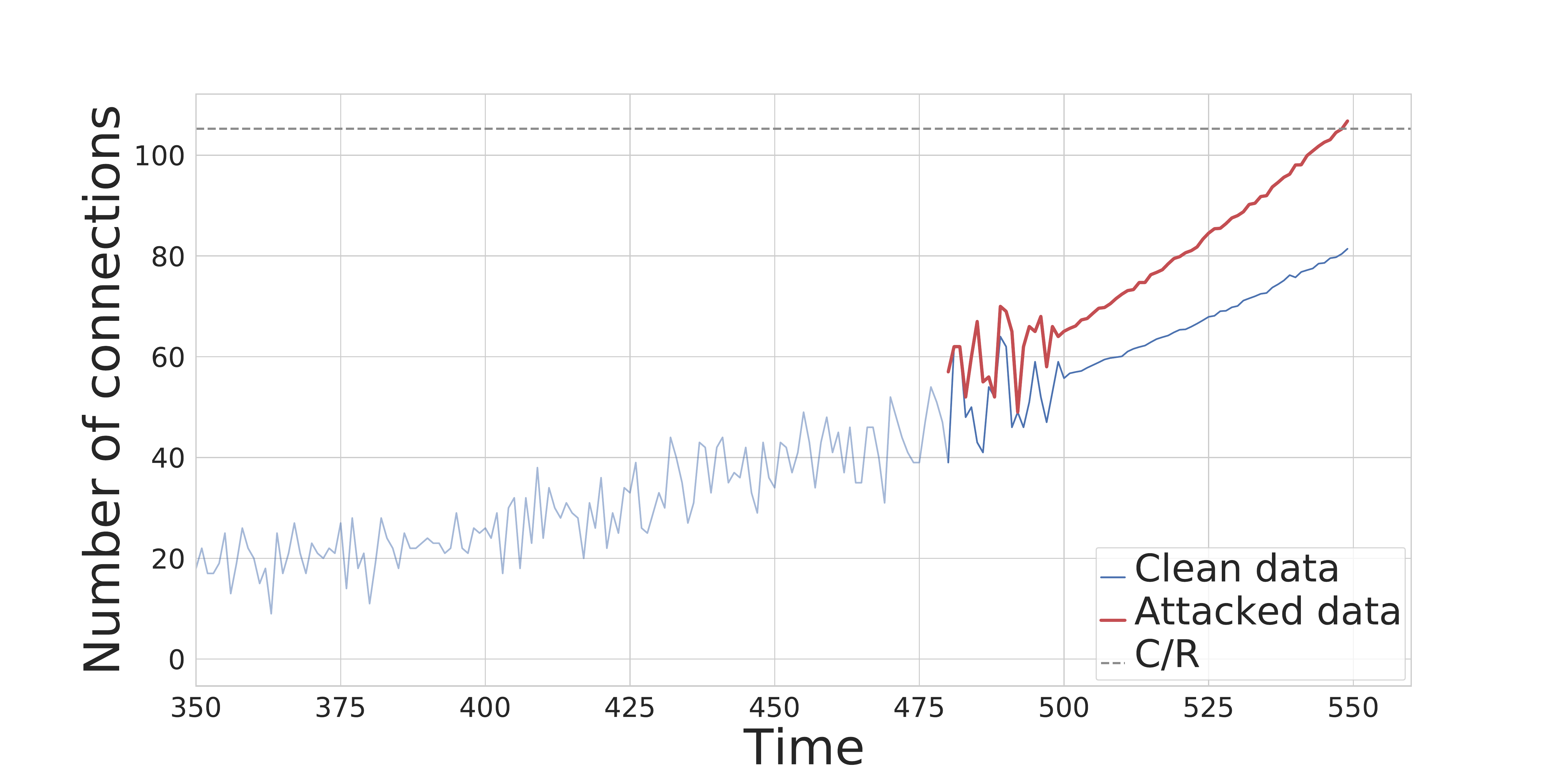}
            \caption[]%
            {Attacked data and forecast for predictive mean.}
            \label{fig:atta}
        \end{subfigure}
        \hfill
        \begin{subfigure}[b]{0.49\textwidth}  
            \centering 
            \includegraphics[width=1.0\textwidth]{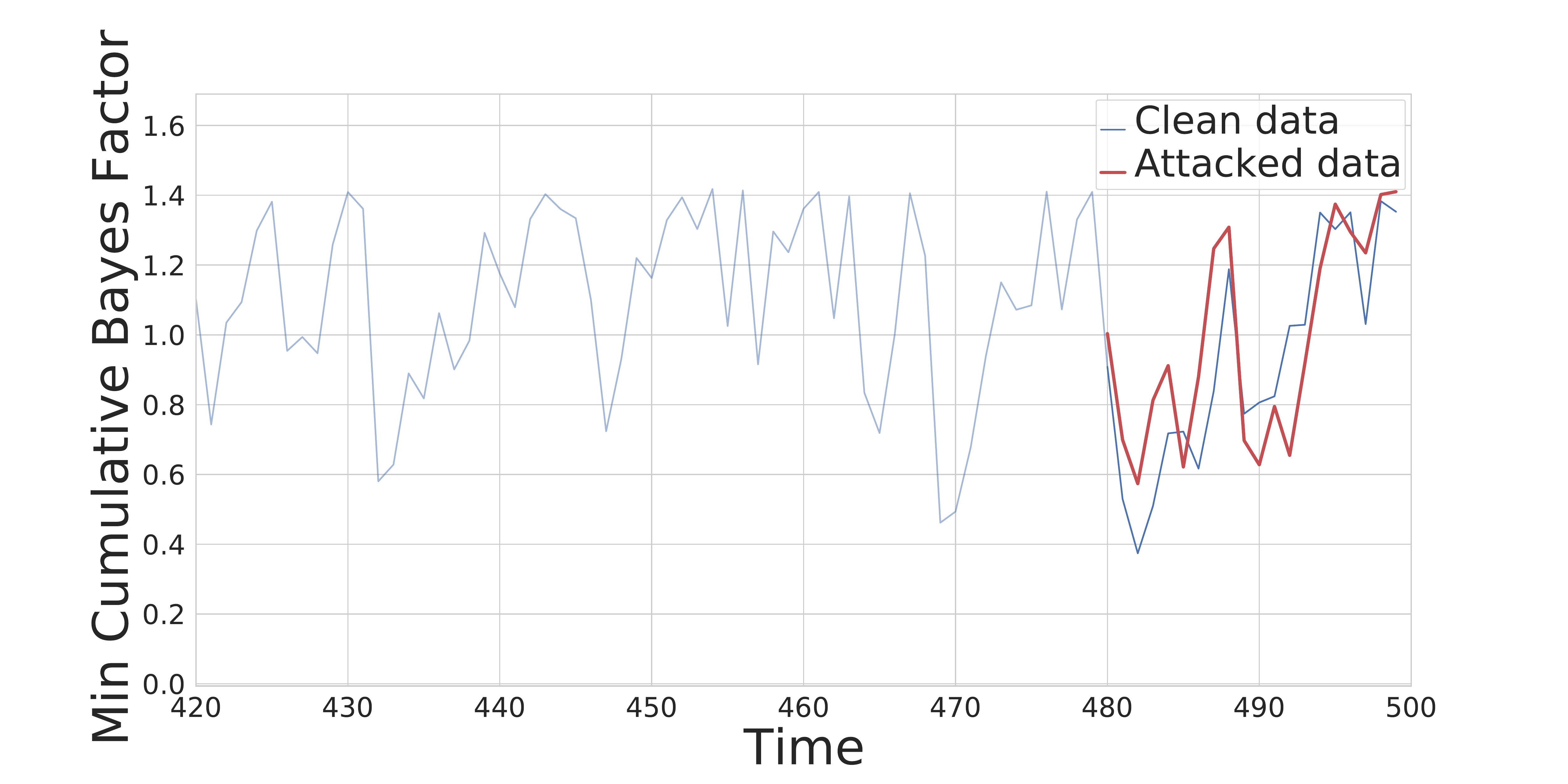}
            \caption[]%
             {Min cumulative Bayes factor, $V_t$. \\ ~}
            \label{fig:attb}
        \end{subfigure}
        \caption[]%
        {Attacked and original data, forecast for predictive mean and $V_t$.} 
        \label{fig:att}
\end{figure}

Continuing with our running example\footnote{The code to reproduce all the experiments in the paper is available at \url{https://github.com/roinaveiro/attacksSSMs}}, imagine that at time $\alpha=500$ the ad company ($D$) must decide about placing an ad on the monitored node, that will appear at a certain time $\beta=550$ in the future, in which the company is interested on. The cost of placing the ad is $C=100$. The reward perceived by the company per user watching the ad is $R=0.95$. Assuming that $D$ is risk neutral, it is straightforward to see that her posterior predictive utility is $R \cdot \mathbb{E}[y_\beta \vert \mathcal{D}_\alpha] - C$ if she decides to place the ad and 0 otherwise.
%
%
$\mathbb{E}[y_\beta \vert \mathcal{D}_\alpha]$ is the posterior predictive mean for the number of connections at time $\beta$. Thus, $D$ should place the ad if $\mathbb{E}[y_\beta \vert \mathcal{D}_\alpha] > \frac{C}{R}$. Having a good estimate of $\mathbb{E}[y_\beta \vert \mathcal{D}_\alpha]$ is crucial to inform $D$'s decision. To that end, $D$ fits a conditionally Poisson model with local linear growth in the latent process \cite{chen2018scalable}. To fit this model, $D$ uses data of previous traffic flow from time $0$ until time $\alpha=500$. The blue line in Figure \ref{fig:atta} shows the original data and the Monte Carlo estimate of the predictive mean from $t=500$ to $t=550$. As can be seen, the predictive mean of the number of connections at time $t=550$ is around $80$, far below $C/R \simeq 105$. Thus, the company should decide not to spend resources on placing the ad. 

Now imagine that $A$, an insider in the company, has resources to create fake connections to the monitored node from time $t=\alpha-h$ (with $h=20$) until time $t=\alpha$. His goal is to make $D$ waste resources in placing the ad (when it is not recommended as we have seen). He can just create fake connections that will be added to the normal traffic flow. $D$ uses a monitoring system as described in Section \ref{sec:poi_bay}. $A$ knows this and, to avoid detection, chooses his attack approximately solving \eqref{eq:opt2}. The red line in Figure \ref{fig:atta} shows that, adding few connections from time $t=480$ to $t=500$, will produce a flip of $D$'s optimal strategy, as the tainted predictive mean at time $t=550$ surpasses $C/R$. In addition, these connections are added in a very subtle way, and they do not trigger an alarm as can be seen in Figure \ref{fig:attb}. We see that the minimum cumulative Bayes factor for the attacking period is not substantially different from that under original data.

\subsection{Optimal inventory problem}

In the running example we have generated poisonous observations targeting a conditionally Poisson model with a local linear growth in the latent process. An interesting aspect of the proposed attacks is that they are applicable to any model, as long as we can compute the predictive distribution of future observations to get the cumulative Bayes factors. 
In this section, we another case study with a slightly different dynamic forecasting model.

A supermarket monitors daily sales of a certain product. It produces predictions of $y_t$, the number of sales in day $t$, using a conditionally Poisson model with local lineal growth and a weekly seasonal component in the latent process. At time $t=\alpha=70$, the supermarket must determine the optimal inventory $d$ of the product for next weekend (including Friday). Let's call $y_{w} = y_{75} + y_{76} + y_{77}$ to the number of sales of that product in the weekend.
If $c$ is the cost per unit of the product, $p$ is the selling price of each unit and $s$ its resale price, the utility perceived by the company when the inventory is $d$ and the demand for the weekend is $y_w$ is
\begin{equation*}
    u(d, y_{w}) = \left \{ \begin{matrix}  (p-c)y_{w} & \mbox{if } & y_{w} > d, 
    \\(p-c)y_{w} + (s-c)(d-y_{w}) & \mbox{if } & y_{w} \leq d\end{matrix}\right. 
\end{equation*}
Where we assume that the company is risk neutral. It is straightforward to see that the posterior predictive utility is
\begin{equation} \label{eq:opt_inv}
    \Psi(d\vert \mathcal{D}_\alpha) = (p-c)\sum_{q=0}^{d-1} p(y_{w} > q \vert \mathcal{D}_\alpha) + (s-c)\sum_{q=0}^{d-1} p(y_{w} \leq q \vert \mathcal{D}_\alpha).
\end{equation}
Optimal inventory can be thus decided maximizing expected utility with respect to $d$. The posterior predictive distributions $p(y_w \vert \mathcal{D}_{\alpha})$ can be estimated using the conditionally Poisson dynamic model.
\begin{figure}[h!] 
    \centering
    \includegraphics[width=0.8\textwidth]{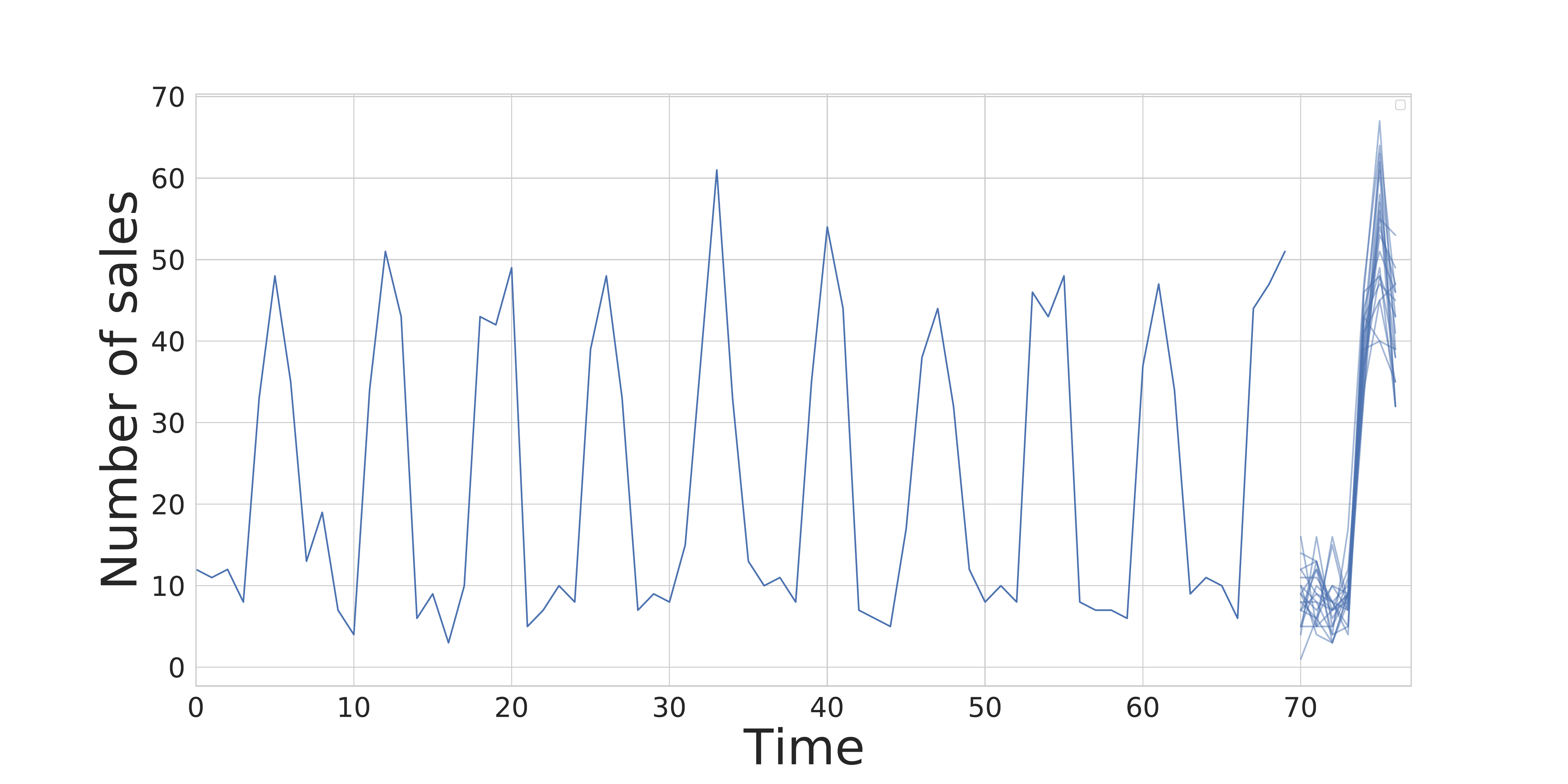}
    \caption{Original data and samples from predictive distribution.}%
    \label{fig:orig_sales}
\end{figure}

Figure \ref{fig:orig_sales} illustrates the daily sales from day 0 until day 70. In addition, we plot several samples from the posterior predictive distribution for days $y_{75}, y_{76}$ and $y_{77}$. Using this data to solve \eqref{eq:opt_inv} the optimal inventory is 136 units.

Now, assume that there is an attacker that has access to the supermarket's monitoring system and can start faking the number of sales three weeks before day $\alpha$. His goal is to make the supermarket buy at least 20\% less units of the product than she should. That means that the optimal inventory must be less or equal than 116. Again, the supermarket uses a monitoring system as the one described in Section 2, and the attacker wants to fake the data in such a way that the monitor is not triggered.

\begin{figure}[h!] 
        \centering
        \captionsetup{justification=centering}
        \begin{subfigure}[b]{0.49\textwidth}
            \centering
            \includegraphics[width=1.0\textwidth]{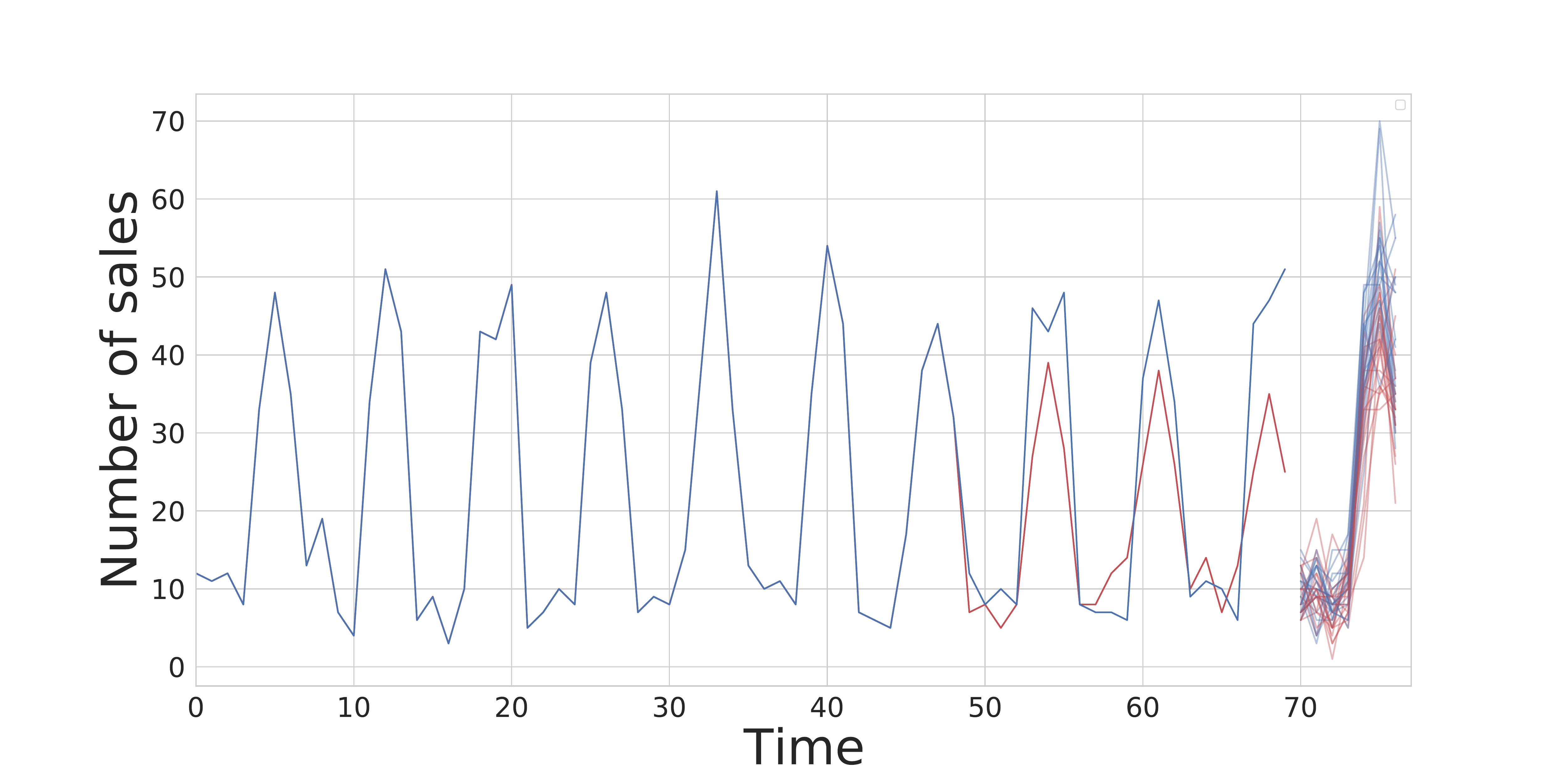}
            \caption[]%
            {Attacked data and samples from predictive distribution.}
            \label{fig:att_inva}
        \end{subfigure}
        \hfill
        \begin{subfigure}[b]{0.49\textwidth}  
            \centering 
            \includegraphics[width=1.0\textwidth]{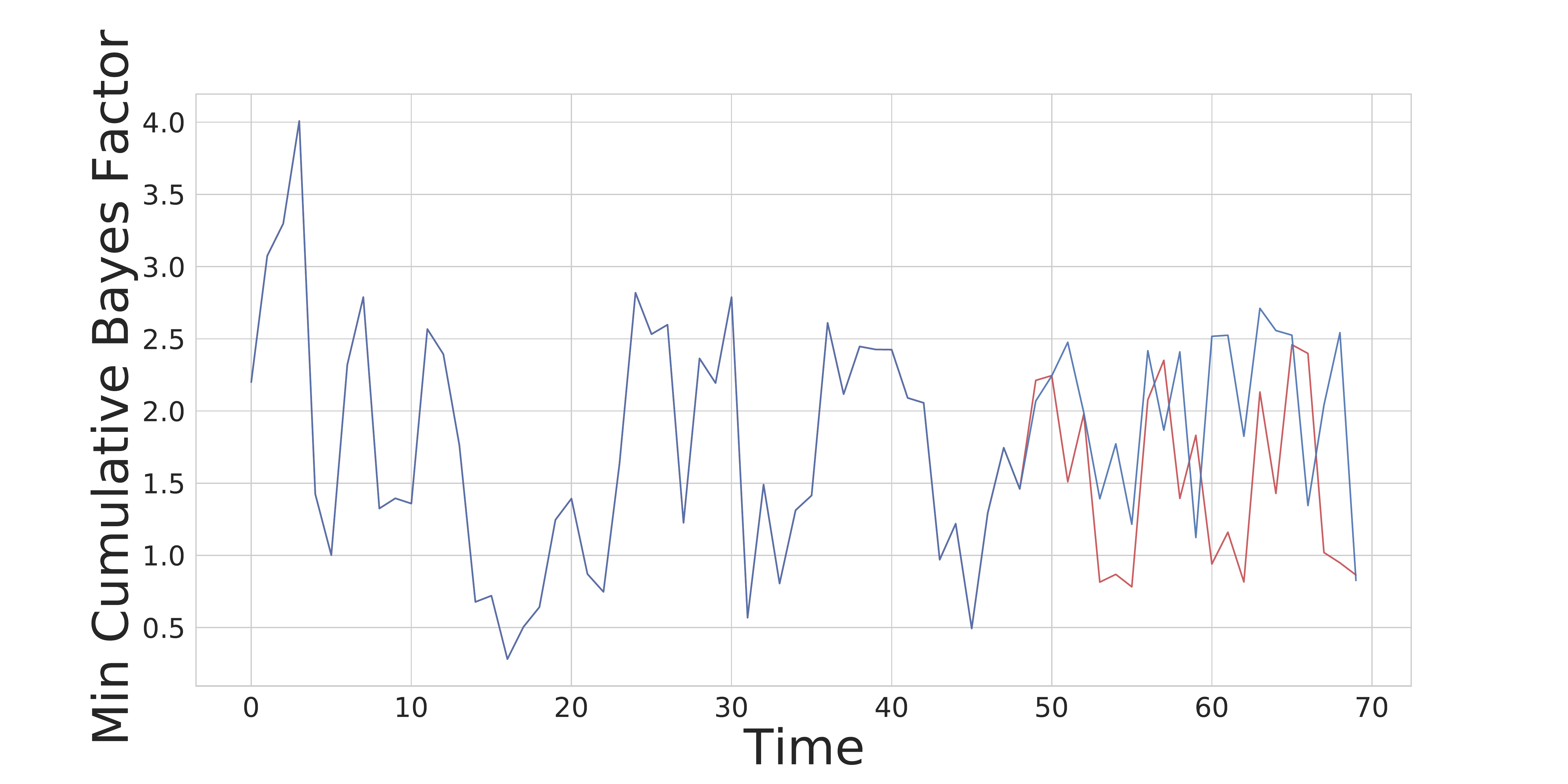}
            \caption[]%
             {Min cumulative Bayes factor, $V_t$. \\ ~}
            \label{fig:att_invb}
        \end{subfigure}
        \caption[]%
        {Results for data attacked solving \eqref{eq:opt2}} 
        \label{fig:att_inv}
\end{figure}

Figure \ref{fig:att_inva} shows data and samples from the posterior predictive distribution under both original (blue line) and attacked data (red line). The attacked data has been computed solving \eqref{eq:opt2}. Under these modified observations, the optimal inventory is 116, thus fulfilling the attacker goal. In addition, Figure \ref{fig:att_invb} shows that minimum cumulative Bayes factor for the attacking period is not substantially different from that under original data. Thus, we can conclude that the attacks will not trigger the monitor.
It is interesting to see that by solving \eqref{eq:opt2}, we learn that, in order to decrease the optimal inventory days 75 through 77 while
not triggering the monitor, the demand in the previous three weekends has to decrease gradually. In this way, the model can adapt to the new low demand scenario, without detecting any aberrant behavior.

\begin{figure}[h!] 
        \centering
        \captionsetup{justification=centering}
        \begin{subfigure}[b]{0.49\textwidth}
            \centering
            \includegraphics[width=1.0\textwidth]{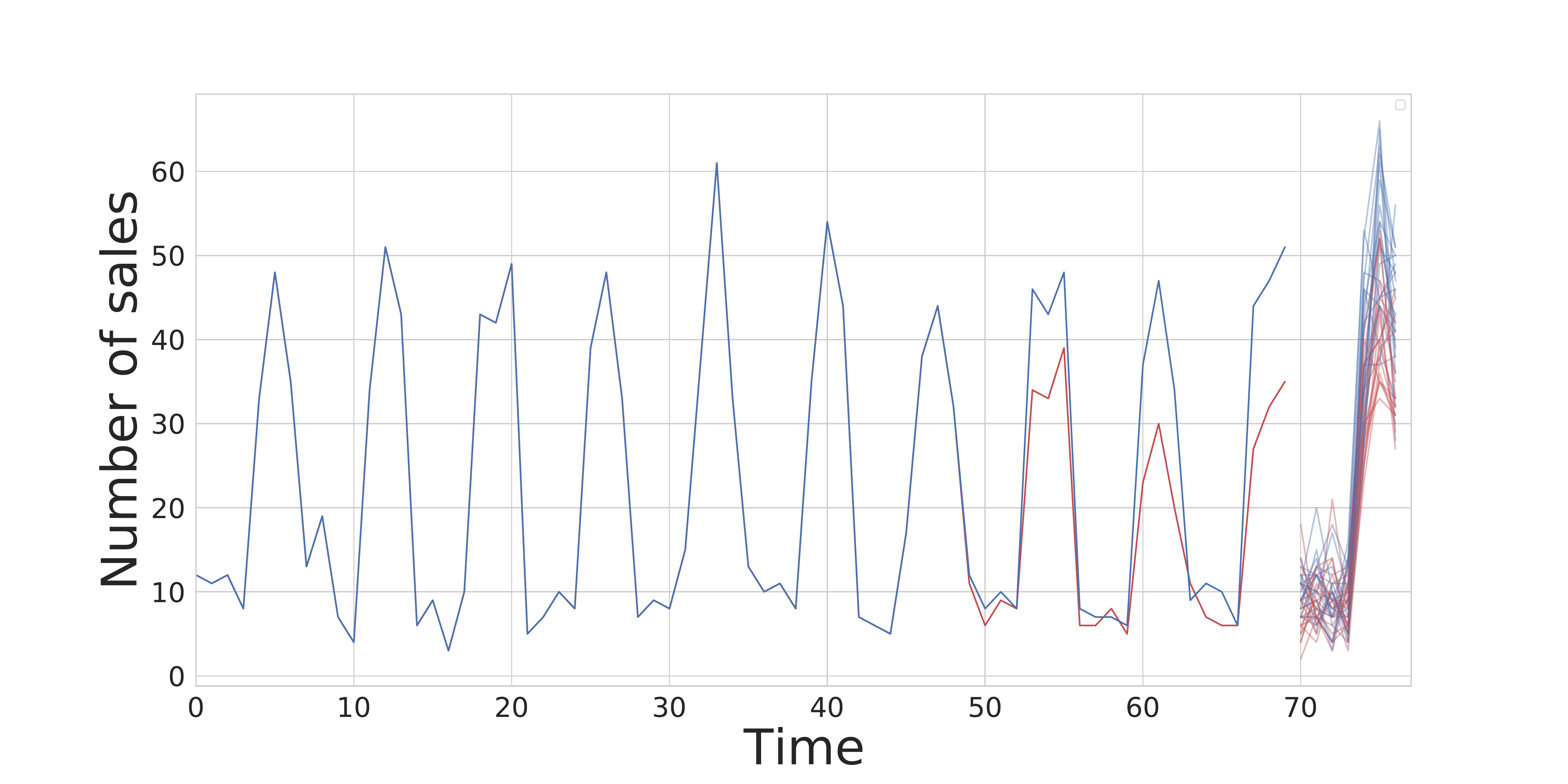}
            \caption[]%
            {Attacked data and samples from predictive distribution.}
            \label{fig:norm_inva}
        \end{subfigure}
        \hfill
        \begin{subfigure}[b]{0.49\textwidth}  
            \centering 
            \includegraphics[width=1.0\textwidth]{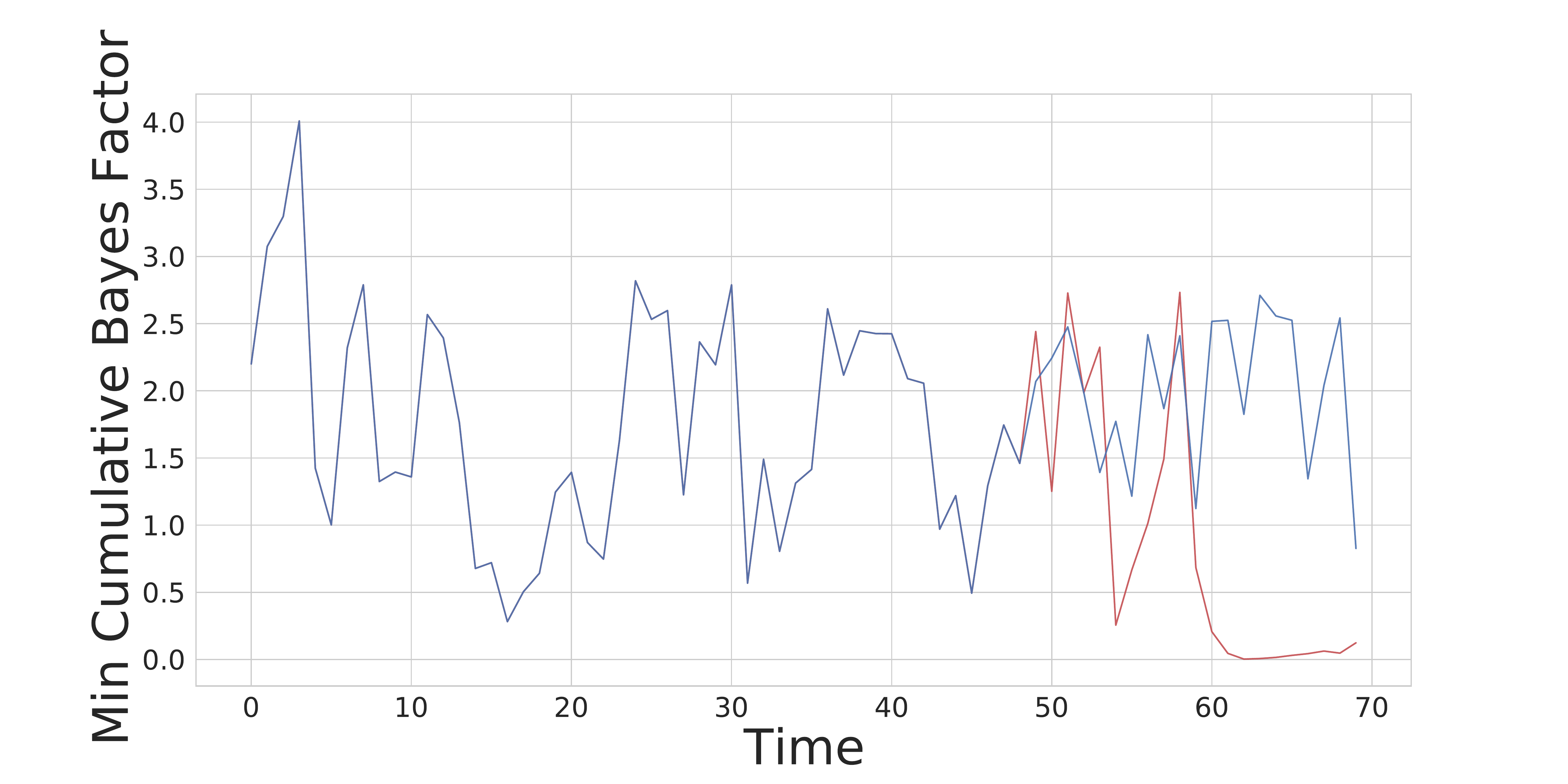}
            \caption[]%
             {Min cumulative Bayes factor, $V_t$. \\ ~}
            \label{fig:norm_invb}
        \end{subfigure}
        \caption[]%
        {Results for data attacked solving \eqref{eq:opt1}} 
        \label{fig:norm_inv}
\end{figure}

Finally, it is of interest to see what happens if we look for the attacked data solving \eqref{eq:opt1} instead of \eqref{eq:opt2}, that is limiting the size of the perturbations measured under certain norm. Figure \eqref{fig:norm_inva} shows data and samples from the posterior predictive distribution under original and attacked data, where the attacked data has been computed solving \eqref{eq:opt1} using the L2 norm. In this case, again the optimal inventory is 116. However, this attack will trigger the monitor as suggested by the minimum cumulative Bayes factors in Figure \eqref{fig:norm_invb}.
Unlike in the previous case, we are not learning to decrease gradually the demand in the previous three weekends and consequently, we cannot find a way to decrease the optimal inventory without triggering the monitor.

\section{Discussion}

We have presented a decision analysis-based approach to generate data manipulation attacks against Bayesian forecasting dynamic models.
Our framework has two central differences with previous approaches: (1) the attacker's goal is to change the decision made by the defender, rather than drive the predictions towards an specific target; (2) the formalization  of ``subtle" attack is based on cumulative Bayes factors rather than the size of the manipulations.

Several further directions of research could be explored. First of all, we have assumed an attacker that has full knowledge about the defender. Considering limited knowledge cases is an interesting way to go. Secondly, we have focused on static attacks: the attacker makes a single data manipulation decision in view of $D$'s model and the data that $D$ will be receiving. A more realistic case would be the dynamic attack: the attacker decides the attack for the next time period, the defender updates the manipulated data point and updates her model, then the attacker decides the next attack, an so on.


\acknowledgements{I would like to thank SAMSI, AXA, the FBBVA, the Trustonomy project and professor Mike West for insightful discussions.}


\bibliographystyle{abbrv}           
\bibliography{references}               





\end{document}